\documentclass[10pt,twocolumn,letterpaper]{article}

 \usepackage[pagenumbers]{cvpr} 

\newcommand{\RQlg}{10.8}
\newcommand{\RQtaming}{9.6}
\newcommand{\RQours}{4.0}

\newcommand{\RQspread}{0.3}

\definecolor{cvprblue}{rgb}{0.21,0.49,0.74}
\usepackage[pagebackref,breaklinks,colorlinks,allcolors=cvprblue]{hyperref}

\def\confName{CVPR}
\def\confYear{2027}

\title{ActPrune: Action-Sensitive Pruning of Gaussian World Models for Robotic Manipulation}

\author{Jingxing Li\\
Arizona State University\\
Tempe, AZ, 85281\\
{\tt\small jingxing@asu.edu}
\and
Yongjae Lee\\
Arizona State University\\
Tempe, AZ, 85281\\
{\tt\small ylee298@asu.edu}
\and
Abhay Kumar Yadav\\
Johns Hopkins University\\
Baltimore, MD, 21218\\
{\tt\small ayadav13@jh.edu}
\and
Rama Chellappa\\
Johns Hopkins University\\
Baltimore, MD, 21218\\
{\tt\small rchella4@jhu.edu}
\and
Deliang Fan\thanks{Corresponding author}\\
Arizona State University\\
Tempe, AZ, 85281\\
{\tt\small dfan@asu.edu}
}

\begin{document}
\maketitle
\begin{abstract}
3D Gaussian Splatting is moving from view synthesis into robotics, where online Gaussian world models let manipulation policies act on the predicted motion of objects.
Fitting and reading tens of thousands of Gaussians at every step is costly, so these models need pruning, but existing pruning criteria were designed for rendering: they rank a Gaussian by its contribution to the image.
A robot is judged by whether its task succeeds, not by how the scene renders.
We show that the rendering quality of a pruned world model does not decide the success of the policy that reads it, and we redesign the pruning criterion around the action.
ActPrune (action-sensitive pruning) scores each Gaussian by the sensitivity of the predicted action to its position and motion coefficients, the only quantities the policy reads, estimated once per episode with two randomized Jacobian projections.
It keeps all robot Gaussians and fills the budget with the highest-scoring others, without retraining.
On PhysMani-Bench we prune \textbf{97\%} of the non-robot Gaussians with ActPrune and, at the same budget, with five existing criteria: opacity and four rendering-based ones.
ActPrune stays on par with the full model (46.3\% against 44.7\% mean success rate over the 16 task variants), whereas the other criteria fall to between 42.8\% and 37.3\%, the latter barely above the 36.6\% of a policy that keeps only the robot.
The criteria that render the pruned scene best (up to \RQlg\,dB PSNR) keep almost none of the task objects, while ActPrune renders it worst (\RQours\,dB).
At this budget, the peak GPU memory of world model and policy falls from 3.6 to 2.2\,GiB and each decision is 1.13$\times$ faster.
\end{abstract}
    
\section{Introduction}
\label{sec:intro}

3D Gaussian Splatting~\cite{kerbl20233dgs} was developed for view synthesis and is increasingly used in robotics, as a scene representation and as a world model for manipulation~\cite{lu2024manigaussian,abouchakra2025pegs,jiang2025phystwin,zhang2025pgnd}.
PhysMani~\cite{yun2026physmani} is a recent example for moving objects, where a policy has to act on where an object will be, not on where it was last observed: a set of 3D Gaussians with a physically constrained velocity field~\cite{li2025freegave} is fitted to the incoming camera frames during the episode, and a 3D policy reads the predicted motion of these Gaussians when it chooses an action.
On PhysMani-Bench its authors report a mean success rate of 45.9\%, against at most 37.8\% for 3D policies~\cite{gkanatsios2025flowmatchactor,ke2024diffuseractor} and 8.3\% for a fine-tuned $\pi_{0.5}$~\cite{pi05}.

This accuracy is paid for at inference time.
The world model is not a feed-forward network: for every new frame it runs 50 optimization iterations on its motion network and 7 on the Gaussian parameters, each rendering about thirty thousand Gaussians, and at every decision the policy searches all of these Gaussians for the neighbors of its visual points.
In our measurements, one decision, which includes ten world-model updates, consumes 2.5\,s of GPU compute, 2.2\,s of it in the world model, and world model and policy together peak at 3.6\,GiB of GPU memory.

\begin{figure*}[t]
  \centering
  \includegraphics[width=\textwidth]{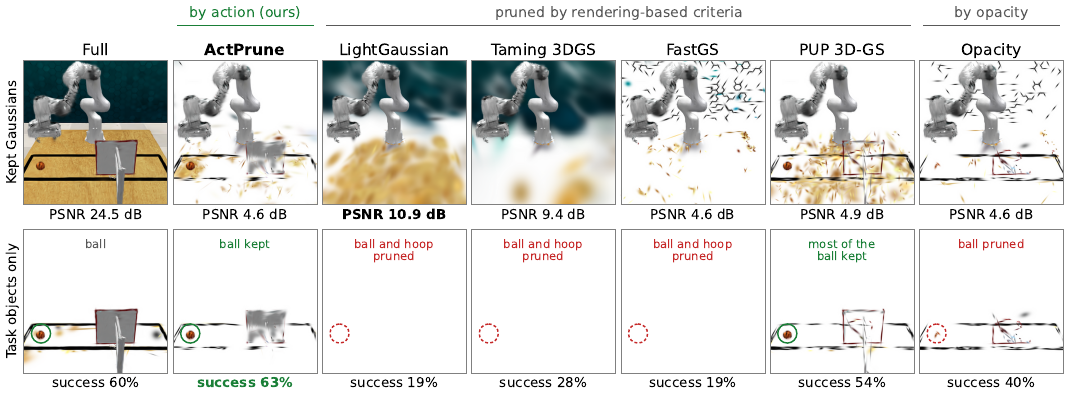}
  \caption{\textbf{Rendering quality does not decide success.} One Drop to Hoop scene pruned by each criterion to all robot Gaussians plus 3\% of the others. Top: all kept Gaussians, with the PSNR of this view against the camera image. Bottom: only the kept Gaussians of the task objects (robot, floor, table, and walls hidden), with the success rate of the policy on this task (80 trials); the circle marks the ball, solid where it is kept and dashed where it is pruned. LightGaussian and Taming 3DGS render the scene best, prune every task object, and succeed in 19\% and 28\% of the trials. PUP 3D-GS, whose score rewards Gaussians whose displacement changes the image, keeps most of the ball and part of the hoop and comes second. ActPrune, tied for the lowest PSNR, keeps the ball, the backboard, and the court lines and succeeds as often as the full model.}
  \label{fig:render-keep}
\end{figure*}

A natural remedy is to keep fewer Gaussians.
Gaussian pruning and densification are well studied~\cite{fan2024lightgaussian,HansonTuPUP3DGS,ren2025fastgs,mallick2024taming}, but their criteria come from view synthesis: they rank a Gaussian by its contribution to the rendered images, because image quality is the goal there.
In robotics the goal is different: the Gaussians are read by a policy, and what counts is whether the task succeeds.
Whether preserving the images also preserves the behavior of the policy has not been examined.
The policy of PhysMani never reads color, opacity, or shape; it reads only the positions and predicted motion of the Gaussians near each of its visual points.
A Gaussian that matters for the image can therefore be irrelevant to the action, while a small moving object that covers few pixels can decide it (Fig.~\ref{fig:render-keep}).
Pruning for a robot thus needs a criterion redesigned around the action instead of the image.

We redesign the criterion in this way, defining it at the policy's output, and call the method ActPrune (action-sensitive pruning).
The path from a Gaussian's position and motion coefficients to the action is differentiable, so we score each Gaussian by the sensitivity of the predicted action to them, estimated once per episode with two randomized Jacobian projections.
ActPrune keeps all robot Gaussians and the highest-scoring others up to a budget.
The other Gaussians are deleted once, the policy stays frozen, and the reduced world model continues to update online for the rest of the episode.

We compare ActPrune with five existing criteria, opacity and four rendering-based ones (LightGaussian~\cite{fan2024lightgaussian}, PUP 3D-GS~\cite{HansonTuPUP3DGS}, FastGS~\cite{ren2025fastgs}, and Taming 3DGS~\cite{mallick2024taming}), at the same budget on all 16 task variants of PhysMani-Bench, and find that the rendering quality of the pruned world model does not decide the success of the policy (Fig.~\ref{fig:render-keep}).
With \textbf{97\%} of the non-robot Gaussians pruned (all robot Gaussians and 3\% of the others kept), LightGaussian and Taming 3DGS render the pruned scene best (\RQlg{} and \RQtaming\,dB PSNR on average), and ActPrune renders it worst (\RQours\,dB), because it removes most of the floor, table, and walls.
Success is ordered differently: ActPrune stays on par with the full model, with a mean success rate of 46.3\% over the 16 task variants against 44.7\%, whereas every other criterion falls below the full model, from 42.8\% (PUP 3D-GS) down to 37.3\% (Taming 3DGS), barely above the 36.6\% of a policy that keeps only the robot.

Where the policy relies on the world model, what decides success is which objects are kept.
The criteria that render best spend the budget on the floor, table, and walls, which cover most pixels, and keep almost none of the objects that the task involves (task objects); opacity and PUP 3D-GS keep a smaller share of them than ActPrune on nearly every task.
The criteria differ most on Drop to Hoop, where the policy relies most on the task objects in the world model: ActPrune keeps on average 34 of the 38 Gaussians of the ball, PUP 3D-GS 18, opacity 9, and LightGaussian, FastGS, and Taming 3DGS none, and success follows: ActPrune succeeds in 50 of 80 trials and the full model in 48, against 43, 32, and 15--22 for the other criteria.
At this budget, the peak GPU memory of world model and policy falls from 3.6 to 2.2\,GiB and each decision is 1.13$\times$ faster.

In summary, we contribute
(i) the finding that the rendering quality of a pruned Gaussian world model does not decide the success of the policy that reads it;
(ii) an analysis of what the policy reads from the world model: only the position and motion of the Gaussians, with an action that depends on the motion at a few locations next to the task objects and not on the appearance that rendering-based criteria preserve; and
(iii) ActPrune, a training-free criterion that prunes the world model by the sensitivity of a frozen policy's action with a single scoring pass per episode; it removes \textbf{97\%} of the non-robot Gaussians while staying on par with the full model, has the highest success rate of all criteria at the same budget while the others fall as low as 37.3\%, and reduces peak GPU memory from 3.6 to 2.2\,GiB and makes each decision 1.13$\times$ faster.

\section{Related Work}
\label{sec:related}

\paragraph{Compact 3D Gaussian Splatting.}
Many methods make 3D Gaussian Splatting~\cite{kerbl20233dgs} more compact.
Pruning and densification criteria score each Gaussian by its contribution to the rendered training views: LightGaussian~\cite{fan2024lightgaussian} by opacity, transmittance, and volume, PUP 3D-GS~\cite{HansonTuPUP3DGS} by the sensitivity of the reconstruction error to its position and scale, Taming 3DGS~\cite{mallick2024taming} by a score that steers densification toward a budget, and FastGS~\cite{ren2025fastgs} by multi-view consistency for densification and pruning.
Mini-Splatting~\cite{fang2024minisplatting} and Speedy-Splat~\cite{hanson2025speedysplat} simplify the set during training, Compact 3DGS~\cite{lee2024compact3dgs}, LP-3DGS~\cite{zhang2024lp3dgs}, and MaskGaussian~\cite{liu2025maskgaussian} learn masks, GaussianSpa~\cite{zhang2025gaussianspa} imposes sparsity in the optimization, and EAGLES~\cite{girish2024eagles} and Niedermayr \etal~\cite{niedermayr2024compressed} quantize per-Gaussian attributes.
For dynamic scenes, Katsumata \etal~\cite{katsumata2024compactdynamic} model positions and rotations as compact functions of time, 4DGS-1K~\cite{yuan2025fourdgs1k} prunes Gaussians with a short temporal span, and MEGA~\cite{zhang2025mega} replaces spherical-harmonic coefficients with a base color and a shared predictor.
All of them are guided by rendered images.
In our setting the consumer is a policy that reads positions and motion rather than pixels, the model is pruned once inside an episode while it continues to be fitted online, and we find that the image quality a criterion preserves does not decide the policy's success.

\paragraph{Dynamic Gaussians and world models for manipulation.}
Gaussians extended over time represent dynamic scenes through per-Gaussian motion with rigidity priors~\cite{luiten2024dynamic3dgs}, deformation fields over a canonical space~\cite{yang2024deformable3dgs,wu2024fourdgs}, or per-frame transformations learned on the fly~\cite{sun2024threedgstream}.
In robotics, ManiGaussian~\cite{lu2024manigaussian} predicts future Gaussian scenes to supervise a manipulation policy, PEGS~\cite{abouchakra2025pegs} attaches Gaussians to simulated particles and corrects them online from images, PhysTwin~\cite{jiang2025phystwin} and PGND~\cite{zhang2025pgnd} build physics-based and learned digital twins rendered with Gaussians, and DreMa~\cite{barcellona2025drema} combines Gaussian world models with a simulator to augment demonstrations.
PhysMani~\cite{yun2026physmani} fits a velocity field~\cite{li2025freegave} to Gaussians online and feeds their predicted motion to a policy built on 3D FlowMatch Actor~\cite{gkanatsios2025flowmatchactor}, which builds on Act3D~\cite{gervet2023act3d} and 3D Diffuser Actor~\cite{ke2024diffuseractor} and is related to 3D policies such as DP3~\cite{ze2024dp3}.
These systems prune, if at all, for reconstruction; we ask which Gaussians the policy needs.

\paragraph{Decision-aware world models.}
In model-based reinforcement learning, a model with lower prediction error does not necessarily give a better policy: the likelihood of learned dynamics need not correlate with control performance~\cite{lambert2020objectivemismatch}.
Value-aware model learning~\cite{farahmand2017valueaware} and the value equivalence principle~\cite{grimm2020valueequivalence} judge a model by its effect on value functions instead of by its prediction error, and VaGraM~\cite{voelcker2022vagram} weights model errors by the gradient of the value function, which is closest in spirit to our score.
These methods change how the model is trained and are defined through a value function.
Our setting has no value function: the policy is trained by imitation, and the world model is fitted online to camera images.
We therefore leave training unchanged and measure the sensitivity of the action directly, to select at inference time the Gaussians that a frozen policy needs.

\paragraph{Efficient policies and sensitivity-based pruning.}
Robot policies are made faster by reducing what they process: VLA-Cache~\cite{xu2025vlacache} reuses visual tokens that change little between frames, and EfficientVLA~\cite{yang2025efficientvla} prunes layers and visual tokens and caches action-head features, following token reduction in vision transformers and vision-language models~\cite{rao2021dynamicvit,bolya2023tome,chen2024fastv}; real-time chunking~\cite{black2025rtc} instead hides latency by executing action chunks asynchronously.
ActPrune removes policy input tokens and world-model primitives at once, because both are the same Gaussians.
Its score belongs to the line of sensitivity-based pruning, from second-order criteria~\cite{lecun1989obd,hassibi1992obs} to first-order Taylor and connection-sensitivity scores~\cite{molchanov2017pruning,molchanov2019importance,lee2019snip}.
Unlike these, it measures the sensitivity of the output action rather than of a training loss, with respect to input primitives rather than weights or channels, and it estimates squared Jacobian norms with random projections in the manner of Hutchinson's trace estimator~\cite{hutchinson1989trace}.
Closest in spirit are task-driven point cloud sampling, which trains a network to select the points that a downstream task needs~\cite{dovrat2019learningtosample,lang2020samplenet}, and point cloud saliency maps, which score each point by its contribution to a recognition loss~\cite{zheng2019pointcloudsaliency}.
ActPrune requires no training, scores with respect to the action of a closed-loop policy, and removes primitives that the world model continues to fit afterwards.

\section{ActPrune: Action-Sensitive Gaussian Pruning}
\label{sec:method}

ActPrune prunes the Gaussians of PhysMani's online world model while its policy stays fixed.
We first summarize the structure of PhysMani that the method relies on (Sec.~\ref{sec:background}) and measure what its policy actually reads from the world model (Sec.~\ref{sec:readout}).
The method then has three parts: a score that measures how much the action depends on each Gaussian (Sec.~\ref{sec:score}), a rule that fills a budget with the highest-scoring Gaussians (Sec.~\ref{sec:selection}), and a single pruning step inside the episode (Sec.~\ref{sec:procedure}).

\subsection{Background: PhysMani}
\label{sec:background}

PhysMani~\cite{yun2026physmani} consists of a world model and a policy that share one set of 3D Gaussians.

\paragraph{World model.}
At the start of an episode, the depth images of four cameras are back-projected into a 3D point cloud, and 25,000 points sampled from it become the centers of the initial Gaussians, which a short 3D Gaussian Splatting optimization~\cite{kerbl20233dgs} refines and densifies to about thirty thousand, $\mathcal{G}=\{g_i\}_{i=1}^{N}$.
A small network predicts six motion coefficients for every Gaussian, three translational and three rotational, which define a divergence-free velocity field following FreeGave~\cite{li2025freegave}.
The model is fitted online: for each new frame it runs 50 iterations that update the motion network and then 7 iterations that update the Gaussian parameters, each rendering the Gaussians into a random subset of the four views and comparing the result with the observed color and depth images.
At a decision, the world model passes the current position and the motion coefficients of every Gaussian to the policy; we write them as one token per Gaussian,
\begin{equation}
  z_i = [x_i;c_i]\in\mathbb{R}^{9},
\end{equation}
where $x_i\in\mathbb{R}^{3}$ is the current position and $c_i\in\mathbb{R}^{6}$ holds the motion coefficients, which describe the predicted motion of the Gaussian.

\paragraph{Policy.}
The policy extends 3D FlowMatch Actor~\cite{gkanatsios2025flowmatchactor}.
Its visual encoder turns the four camera images into $M=4{,}096$ visual points; visual point $j$ has a 3D position $p_j\in\mathbb{R}^{3}$ and a feature vector $f_j$.
A fusion module connects the policy to the world model.
For visual point $j$, let $\mathcal{N}(j)$ be the set of Gaussians nearest to $p_j$ (the 16 nearest within a fixed radius in the released implementation).
The module starts from a learnable query vector $u$, shared by all visual points, and updates it with three cross-attention blocks that attend to these Gaussians through their positions relative to $p_j$ and their motion coefficients.
Writing the three blocks as $\phi$, the result is one dynamics token
\begin{equation}
  d_j=\phi\big(u,\{(x_i-p_j,\,c_i)\}_{i\in\mathcal{N}(j)}\big),
  \label{eq:readout}
\end{equation}
which is added to $f_j$.
A rectified-flow action head with ten sampling steps then predicts the action from these points, the instruction, and the gripper history, collected as $o_t$:
\begin{equation}
  a_t=\pi_\theta\big(o_t,d_1,\ldots,d_M;\epsilon\big)\in\mathbb{R}^{8},
\end{equation}
where $\epsilon$ is the sampling noise of the action head and $a_t$ comprises position, a quaternion, and gripper openness (the network represents rotation in 6D internally and returns a quaternion).
The world model therefore influences the action only through the $M$ dynamics tokens.

\paragraph{What this structure implies for pruning.}
Every Gaussian is rendered in each of the 57 fitting iterations per frame, and every Gaussian enters the neighbor search of the fusion module at each decision.
Deleting a Gaussian therefore removes work from both modules, and the shared set $\mathcal{G}$ is what we prune.
The path from a token to the action, $z_i\rightarrow d_j\rightarrow a_t$, is differentiable, so the importance of a Gaussian can be measured at the action instead of at the rendered image.

\subsection{What the Policy Reads from the World Model}
\label{sec:readout}

We measure four properties of this readout on the unpruned model; they motivate ActPrune.

\paragraph{Only position and motion are read.}
Each Gaussian contributes its position and six motion coefficients.
Color, opacity, and shape, which determine rendering quality, never reach the policy.

\paragraph{Many Gaussians are not read at all.}
At the first world-model query, 43--46\% of the Gaussians are not among the neighbors of any visual point; they have no effect on the current action.

\paragraph{Static Gaussians change the tokens little.}
A visual point with no Gaussian within the radius still receives a token, a constant vector $d^{\varnothing}$ of norm 7.2.
Over the scene, the tokens of the full model differ from this constant by a norm of only 1.7 on average, because most Gaussians do not move.

\paragraph{The action depends on a few tokens.}
Measured by the gradient of the action with respect to each dynamics token, 12--22 of the 4,096 visual points, those next to the task object, carry 48--64\% of the sensitivity.

\begin{figure*}[t]
  \centering
  \includegraphics[width=\textwidth]{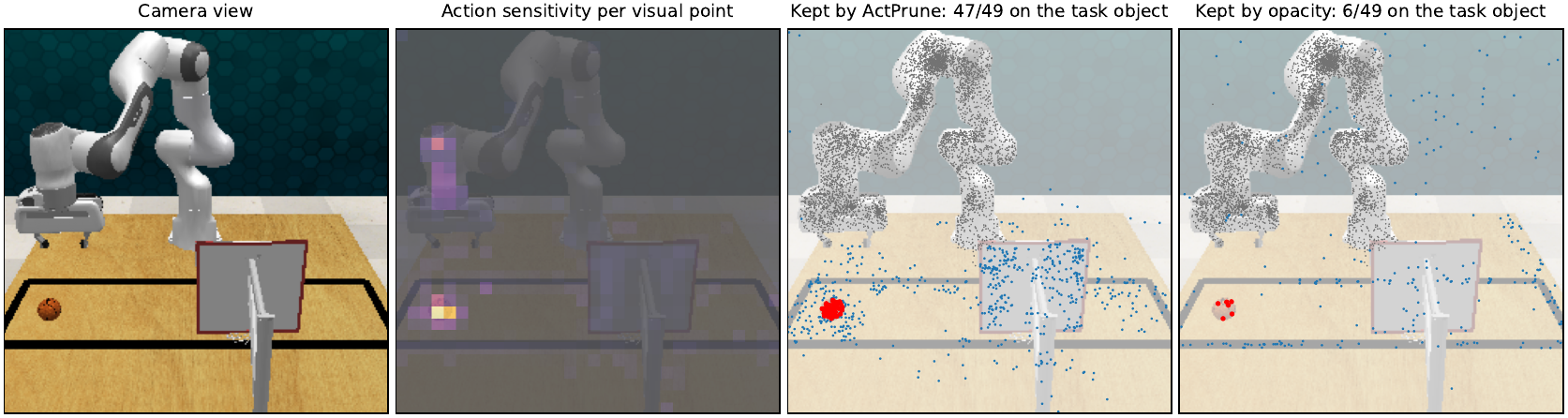}
  \caption{How the policy reads the world model, on one Drop to Hoop state at the first world-model query. From left: camera view; sensitivity of the action to the dynamics token of each visual point (the brightest cell holds 19\% of the total); Gaussians kept by ActPrune and by opacity with 3\% of the non-robot Gaussians (gray: robot, blue: other, red: the ball). In this state the ball has 49 Gaussians; ActPrune keeps 47 of them, opacity 6.}
  \label{fig:readout}
\end{figure*}

Of about thirty thousand Gaussians, the action is thus most sensitive to the motion read at a handful of locations (Fig.~\ref{fig:readout}), and a criterion based on image reconstruction ranks Gaussians by a quantity the policy never reads.

\subsection{Action-Sensitivity Score}
\label{sec:score}

\paragraph{Definition.}
A Gaussian matters to the policy if changing what it reports changes the action.
We score Gaussian $i$ by the squared Frobenius norm~\cite{horn2012matrix} of the Jacobian of the action with respect to its token:
\begin{equation}
  s_i=\left\lVert\frac{\partial a_t}{\partial z_i}\right\rVert_{\mathrm F}^{2},
  \qquad
  \frac{\partial a_t}{\partial z_i}
  =\sum_{j:\,i\in\mathcal{N}(j)}\frac{\partial a_t}{\partial d_j}\,\frac{\partial d_j}{\partial z_i}.
  \label{eq:score}
\end{equation}
To first order, this norm is proportional to the expected squared change of the action under an isotropic perturbation of the token, and it can be estimated with random projections (Eq.~\ref{eq:projection}).
The chain rule on the right shows what the score rewards.
A Gaussian scores high only if some visual point reads it ($i\in\mathcal{N}(j)$), the action is sensitive to that point's dynamics token ($\partial a_t/\partial d_j$), and that token responds to this Gaussian ($\partial d_j/\partial z_i$).
A Gaussian that no visual point reads scores exactly zero, and appearance plays no role.

\paragraph{Estimation.}
The action has eight coordinates, so the exact score would need eight backward passes.
We instead use random projections.
For independent $v_p\sim\mathcal{N}(0,I_8)$,
\begin{equation}
  \widehat{s}_i = \frac{1}{P}\sum_{p=1}^{P}
  \left\lVert\frac{\partial\big(v_p^{\mathsf T}a_t\big)}{\partial z_i}\right\rVert_2^2,
  \qquad P=2,
  \label{eq:projection}
\end{equation}
which satisfies $\mathbb{E}[\widehat{s}_i]=s_i$ because $\mathbb{E}[v_pv_p^{\mathsf T}]=I_8$, as in Hutchinson's trace estimator~\cite{hutchinson1989trace}.
One forward pass of the policy and two backward passes give the scores of all $N$ Gaussians; this takes 0.45\,s per episode in our implementation.
On five saved states, the Gaussians selected with $P=2$ overlap by 82--95\% with those selected with $P=16$ at a budget of 3\% of the non-robot Gaussians.

\paragraph{Settings.}
During scoring we fix the sampling noise $\epsilon$, so that the derivative reflects the input rather than the noise.
We also run the action head with five sampling steps instead of the ten used for acting, which halves the cost of the backward passes.

\subsection{Selection Under a Budget}
\label{sec:selection}

Let $\mathcal{R}$ be the Gaussians with a robot semantic label and $R=|\mathcal{R}|$.
Given a total budget $B\geq R$, we keep every robot Gaussian and fill the rest of the budget with the highest-scoring other Gaussians:
\begin{equation}
  \mathcal{K}_B=\mathcal{R}\;\cup\;\operatorname{top}_{B-R}\big\{\widehat{s}_i : i\notin\mathcal{R}\big\},
  \label{eq:selection}
\end{equation}
where $\operatorname{top}_{k}$ returns the $k$ Gaussians with the highest scores.
We protect the robot because the world model estimates the end-effector pose from the robot Gaussians.

\paragraph{Budget.}
We set the budget per scene from a share $q$ of the non-robot Gaussians, $B=R+\operatorname{round}\big(q\,(N-R)\big)$, so that every criterion keeps the same number of Gaussians in each scene.
The main comparison uses $q=3\%$, about 5,500 of the roughly 31,000 Gaussians of a scene.

\subsection{Pruning Within an Episode}
\label{sec:procedure}

Pruning happens once per episode, at the first decision that reads the world model, which in the evaluated system is the second decision of the episode.
There we compute $\widehat{s}_i$ for every Gaussian on the current observation (Eq.~\ref{eq:projection}), form $\mathcal{K}_B$ (Eq.~\ref{eq:selection}), and delete every other Gaussian.
For the rest of the episode the world model fits and predicts with the remaining Gaussians only, and the policy reads only those.
The motion network and the number of fitting iterations are unchanged, and the cost of scoring and deletion is included in the reported latency.

\subsection{What the Score Measures}
\label{sec:scope}

The score measures how strongly the action at the pruning step depends on each Gaussian.
On eight saved states, more than 99\% of it comes from the six motion coefficients rather than from the position: a Gaussian scores high when the action is sensitive to its motion, whether or not it is currently moving.
On fixed states, deleting high-scoring Gaussians changes the action more than deleting low-scoring ones (supplementary material, Sec.~D).
The effect of deletion over a whole episode is evaluated in closed loop (Sec.~\ref{sec:main-results}), and Sec.~\ref{sec:token-analysis} examines what the policy reads after pruning.

\section{Experiments}
\label{sec:experiments}

\subsection{Evaluation Protocol}

\paragraph{System and tasks.}
We evaluate a frozen PhysMani policy and its online world model on PhysMani-Bench~\cite{yun2026physmani}: eight tasks, each with a normal-speed and a high-speed variant, and 40 official test cases per variant; a trial is one closed-loop run of a case.
World-model initialization, the fitting schedule (50 network and 7 Gaussian iterations per frame), and the ten rectified-flow sampling steps of the policy are those of the released system; the simulator advances ten steps per decision, and the world model is updated at each.

\paragraph{Comparisons.}
Every pruned model keeps all robot Gaussians and the same number of Gaussians in total as ActPrune, so only the ranking differs, and no pruned model is fine-tuned.
We report successes and, for two methods, the paired counts of cases solved by only one of them, with a two-sided exact sign (McNemar) test; $p$-values are unadjusted.

\paragraph{Image quality.}
To measure what the rendering-based criteria are designed to preserve, we replay recorded test episodes to the pruning step, select with every criterion at the budget of the main comparison, and render each kept set from the four cameras of the world model, with no fitting after the deletion.
We report PSNR against the camera images of that frame.

\subsection{Main Results}
\label{sec:main-results}

Table~\ref{tab:main} compares the criteria on every PhysMani-Bench task variant.
The budget is set per scene: every method keeps all robot Gaussians and 3\% of the other Gaussians, ranked by its own score, which leaves about 5,500 of the roughly 31,000 Gaussians of a scene.
Besides opacity we compare with four rendering-based criteria: LightGaussian~\cite{fan2024lightgaussian}, PUP 3D-GS~\cite{HansonTuPUP3DGS}, FastGS~\cite{ren2025fastgs}, and Taming 3DGS~\cite{mallick2024taming}.
Each rendering-based score follows its released implementation and is computed at the pruning step from the four cameras of the world model: LightGaussian sums opacity times transmittance over the pixels a Gaussian contributes to and weights it by volume; PUP 3D-GS takes the log-determinant of the Fisher matrix of the rendered pixels with respect to a Gaussian's position and scale; FastGS and Taming 3DGS provide scores that their authors use during training, which we use here as rankings.

\begin{table*}[t]
\centering
\small
\setlength{\tabcolsep}{7pt}
\begin{tabular}{llrrrrrrr}
\hline
Task & Speed & Full & ActPrune & Opacity & PUP 3D-GS & LightGaussian & FastGS & Taming \\
\hline
Drop to Hoop & Normal & 29/40 & 33/40 & 22/40 & 28/40 & 12/40 & 11/40 & 15/40 \\
Drop to Hoop & High & 19/40 & 17/40 & 10/40 & 15/40 & 3/40 & 4/40 & 7/40 \\
Pick Moving Target & Normal & 32/40 & 32/40 & 31/40 & 32/40 & 33/40 & 33/40 & 25/40 \\
Pick Moving Target & High & 7/40 & 7/40 & 10/40 & 6/40 & 11/40 & 4/40 & 10/40 \\
Insert onto Peg & Normal & 17/40 & 17/40 & 17/40 & 21/40 & 22/40 & 20/40 & 18/40 \\
Insert onto Peg & High & 7/40 & 12/40 & 10/40 & 9/40 & 8/40 & 7/40 & 2/40 \\
Deposit Rubbish & Normal & 25/40 & 26/40 & 21/40 & 21/40 & 22/40 & 23/40 & 21/40 \\
Deposit Rubbish & High & 20/40 & 19/40 & 14/40 & 12/40 & 15/40 & 15/40 & 6/40 \\
Push Button & Normal & 20/40 & 19/40 & 16/40 & 18/40 & 21/40 & 20/40 & 21/40 \\
Push Button & High & 5/40 & 5/40 & 6/40 & 7/40 & 5/40 & 6/40 & 6/40 \\
Beat Buzz & Normal & 29/40 & 29/40 & 30/40 & 28/40 & 31/40 & 29/40 & 31/40 \\
Beat Buzz & High & 14/40 & 22/40 & 22/40 & 21/40 & 22/40 & 21/40 & 23/40 \\
Remove Cup & Normal & 25/40 & 23/40 & 23/40 & 21/40 & 23/40 & 24/40 & 23/40 \\
Remove Cup & High & 19/40 & 22/40 & 20/40 & 19/40 & 20/40 & 19/40 & 17/40 \\
Place Cups & Normal & 11/40 & 8/40 & 11/40 & 11/40 & 11/40 & 9/40 & 9/40 \\
Place Cups & High & 7/40 & 5/40 & 8/40 & 5/40 & 5/40 & 6/40 & 5/40 \\
\hline
\multicolumn{2}{l}{All variants} & 286/640 & 296/640 & 271/640 & 274/640 & 264/640 & 251/640 & 239/640 \\
\multicolumn{2}{l}{Mean SR (\%)} & 44.7 & 46.3 & 42.3 & 42.8 & 41.3 & 39.2 & 37.3 \\
\hline
\end{tabular}
\caption{Successes on PhysMani-Bench when every method keeps all robot Gaussians plus 3\% of the other Gaussians, chosen by its own criterion (Full keeps every Gaussian); one seed over the 40 official test cases of each task variant. No method is fine-tuned after pruning. Mean SR is the success rate averaged over the 16 variants.}
\label{tab:main}
\end{table*}

Over the 16 variants, ActPrune has a mean success rate of 46.3\%, on par with the full model (44.7\%) and ahead of PUP 3D-GS (42.8\%), opacity pruning (42.3\%), LightGaussian (41.3\%), FastGS (39.2\%), and Taming 3DGS (37.3\%).
Paired on the same 640 cases, 88 are solved only by ActPrune and 63 only by opacity ($p=0.050$); the counts are 82 and 60 against PUP 3D-GS ($p=0.078$), 93 and 61 against LightGaussian ($p=0.012$), 100 and 55 against FastGS, 122 and 65 against Taming 3DGS (both $p<0.001$), and 66 and 56 against the full model ($p=0.42$, not a significant difference).
The criteria differ most on Drop to Hoop, where ActPrune succeeds in 50/80 trials, the full model in 48/80, PUP 3D-GS in 43/80, opacity in 32/80, Taming 3DGS in 22/80, and LightGaussian and FastGS in 15/80 each, with paired counts for ActPrune of 24 and 6 against opacity ($p=0.001$), 16 and 9 against PUP 3D-GS ($p=0.23$), and at least 35 against at most 7 for the other three.
On the other 14 variants the differences are smaller.

\subsection{Rendering Quality Does Not Decide Success}
\label{sec:render-success}

Table~\ref{tab:render-success} sets the image quality that each criterion preserves at the budget of Table~\ref{tab:main} next to its successes, and Fig.~\ref{fig:render-keep} shows one scene; per-task values and further scenes are in Sec.~C of the supplementary material.

\begin{table}[t]
\centering
\small
\setlength{\tabcolsep}{5pt}
\begin{tabular}{lrrr}
\hline
 & PSNR & Objects & \\
Criterion & (dB) & kept (\%) & Successes \\
\hline
Full & 27.2 & 100 & 286/640 \\
\hline
LightGaussian & 10.8 & 0 & 264/640 \\
Taming 3DGS & 9.6 & 0 & 239/640 \\
Opacity & 4.3 & 15 & 271/640 \\
FastGS & 4.2 & 0 & 251/640 \\
PUP 3D-GS & 4.1 & 17 & 274/640 \\
ActPrune & 4.0 & 37 & 296/640 \\
\hline
\end{tabular}
\caption{PSNR of the whole image right after pruning, share of task-object Gaussians kept, and successes at the budget of Table~\ref{tab:main}, ordered by PSNR. PSNR and the share are means over 48 replayed test episodes of all task variants (four cameras each).}
\label{tab:render-success}
\end{table}

\paragraph{The criteria that render best do not succeed most.}
LightGaussian and Taming 3DGS preserve the images best, \RQlg{} and \RQtaming\,dB against about 4\,dB for the other criteria, because they spend the budget on the floor, table, and walls, which cover most pixels.
This does not help the policy: on Drop to Hoop LightGaussian and Taming 3DGS succeed in 15 and 22 of 80 trials, against 48 for the full model.
ActPrune has the lowest PSNR, \RQours\,dB, because it removes most of the background and the white background of the renderer shows through, and it has the most successes, both over all variants and on Drop to Hoop.
Opacity, PUP 3D-GS, and FastGS render within \RQspread\,dB of ActPrune, yet on Drop to Hoop opacity and PUP 3D-GS succeed in 32 and 43 of 80 trials, against 50 for ActPrune.
Random selection at the same budget renders Drop to Hoop better than opacity, PUP 3D-GS, and ActPrune (5.5 against 4.4, 4.1, and 3.9\,dB) and succeeds in 28/80 trials, fewer than each of them.
The policy never reads color, opacity, or shape (Sec.~\ref{sec:readout}), so the image is not the quantity it depends on.

\paragraph{Where the policy relies on the world model, what is kept decides success.}
\label{sec:object-keep}
Every trial records the semantic label of each Gaussian before and after pruning, so we can count how many Gaussians of the task objects, those that belong to neither the robot nor the floor, table, and walls, each criterion keeps (Table~\ref{tab:object-keep}).
LightGaussian, FastGS, and Taming 3DGS keep at most about 2\% of the task-object Gaussians in every task measured.
Opacity and PUP 3D-GS keep more, and ActPrune keeps the largest share of the task objects in seven of the eight tasks: more than opacity in all but Push Button and more than PUP 3D-GS in all eight.
PUP 3D-GS keeps more of them than the other rendering-based criteria because its score is the sensitivity of the image to the position and scale of a Gaussian, which is high where moving a Gaussian changes the image, as at the outline of a small object against the floor; LightGaussian and Taming 3DGS instead favor Gaussians that cover many pixels.
Its score still ignores motion and the action: on Drop to Hoop it keeps on average 18 of the 38 Gaussians of the ball, against 34 for ActPrune (Table~\ref{tab:ball}).
LightGaussian, which keeps almost none of them, has 33 fewer successes than the full model on Drop to Hoop and at most eight fewer on any other task; accordingly, the criteria differ most on Drop to Hoop.

\begin{table}[t]
\centering
\small
\setlength{\tabcolsep}{2.6pt}
\begin{tabular}{lrrrrrr}
\hline
 & \multicolumn{6}{c}{Task-object Gaussians kept (\%)} \\
Task & ActPrune & Opac. & PUP & LG & FastGS & Tam. \\
\hline
Drop to Hoop & 13 & 9 & 9 & 0 & 0 & 0 \\
Pick Moving Target & 30 & 25 & 19 & 0 & 1 & 0 \\
Insert onto Peg & 50 & 15 & 19 & 0 & 1 & 0 \\
Deposit Rubbish & 32 & 16 & 13 & 0 & 2 & 0 \\
Push Button & 18 & 22 & 17 & 0 & 2 & 0 \\
Beat Buzz & 42 & 6 & 4 & 0 & 0 & 0 \\
Remove Cup & 64 & 14 & 31 & 0 & 0 & 0 \\
Place Cups & 39 & 10 & 19 & 0 & 0 & 0 \\
\hline
\end{tabular}
\caption{What each criterion keeps in the trials of Table~\ref{tab:main}, both speeds of a task pooled: the share of the task-object Gaussians, those that belong to neither the robot nor the floor, table, and walls. LG: LightGaussian; Tam.: Taming 3DGS.}
\label{tab:object-keep}
\end{table}

On Drop to Hoop the ball is represented on average by 38 of the roughly 32,000 Gaussians of a scene.
Table~\ref{tab:ball} lists how many of them each criterion keeps: successes follow the ball, and the three criteria that keep none of it succeed about as often as the robot alone.

\begin{table}[t]
\centering
\small
\setlength{\tabcolsep}{5pt}
\begin{tabular}{lrrr}
\hline
 & \multicolumn{2}{c}{Gaussians kept} & \\
Criterion & Ball (of 38) & Hoop (of 1,980) & Successes \\
\hline
Full & 38 & 1,980 & 48/80 \\
\hline
ActPrune & 34 & 242 & 50/80 \\
PUP 3D-GS & 18 & 90 & 43/80 \\
Opacity & 9 & 48 & 32/80 \\
Taming 3DGS & 0 & 0 & 22/80 \\
LightGaussian & 0 & 1 & 15/80 \\
FastGS & 0 & 1 & 15/80 \\
\hline
Robot only & 0 & 0 & 21/80 \\
\hline
\end{tabular}
\caption{Drop to Hoop at the budget of Table~\ref{tab:main}: mean number of Gaussians of the ball and of the hoop that each criterion keeps, and successes. Robot only keeps no other Gaussian.}
\label{tab:ball}
\end{table}

\subsection{Other Budgets}
\label{sec:budget-curve}

\begin{figure}[t]
  \centering
  \includegraphics[width=0.8\columnwidth]{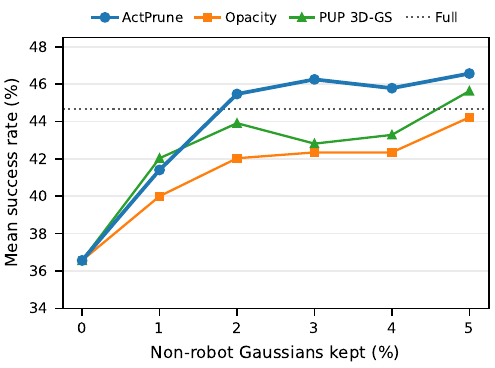}
  \caption{Mean success rate over the 16 task variants versus the share of non-robot Gaussians kept (640 cases per point, the seed of Table~\ref{tab:main}). At 0\% only the robot is kept. Dotted: the full model.}
  \label{fig:budget-curve}
\end{figure}

Table~\ref{tab:budget-mean} and Fig.~\ref{fig:budget-curve} vary the share of non-robot Gaussians for ActPrune, opacity, and PUP 3D-GS on all 16 variants, with the seed and cases of Table~\ref{tab:main} (per-variant tables: supplementary material, Sec.~A).
From 2\% on, ActPrune stays level with the full model and has the highest success rate of the three.
At 1\%, ActPrune, opacity, and PUP 3D-GS are close and below the full model.
LightGaussian, FastGS, and Taming 3DGS stay below opacity and the full model at 3\% (Table~\ref{tab:main}) and at 5\% (supplementary material, Sec.~A), and ActPrune succeeds more often than each of them at both shares (paired $p\leq0.020$).

\begin{table}[t]
\centering
\small
\setlength{\tabcolsep}{6pt}
\begin{tabular}{lrrrrr}
\hline
 & \multicolumn{5}{c}{Share of non-robot Gaussians kept} \\
Criterion & 1\% & 2\% & 3\% & 4\% & 5\% \\
\hline
ActPrune & 41.4 & \textbf{45.5} & \textbf{46.3} & \textbf{45.8} & \textbf{46.6} \\
Opacity & 40.0 & 42.0 & 42.3 & 42.3 & 44.2 \\
PUP 3D-GS & \textbf{42.0} & 43.9 & 42.8 & 43.3 & 45.6 \\
\hline
\end{tabular}
\caption{Mean success rate (\%) over the 16 task variants at each share (640 cases per entry, the seed and cases of Table~\ref{tab:main}; best criterion per share in bold). The full model reaches 44.7\% and the robot alone 36.6\%.}
\label{tab:budget-mean}
\end{table}

\subsection{Memory and Latency}
\label{sec:deploy-latency}

Table~\ref{tab:latency-deploy} reports latency and peak memory at the budget of Table~\ref{tab:main}.
Peak memory falls from 3,737 to 2,224\,MiB, almost entirely in the policy, which searches every Gaussian (2,809 to 1,384\,MiB).
A world-model update takes 162 instead of 199\,ms, and compute per decision falls from 2,470 to 2,183\,ms (1.13$\times$), including the one-time score of 0.45\,s per episode.
The other criteria keep the same number of Gaussians and have similar latency and peak memory after pruning.

Pruning removes the part of an update that grows with the number of Gaussians; the rest, such as the image losses and the updates of the motion network in the 57 fitting iterations per frame, does not depend on it.

\IfFileExists{exp_latency/latency_profile_table.tex}{\begin{table}[t]
\centering
\small
\setlength{\tabcolsep}{3pt}
\begin{tabular}{lrrrrrr}
\hline
 & \multicolumn{3}{c}{Latency (ms)} & \multicolumn{3}{c}{Peak memory (MiB)} \\
Method & WM & Policy & Total & WM & Policy & Total \\
\hline
Full & 2,183 & 287 & 2,470 & 992 & 2,809 & 3,737 \\
ActPrune & 1,860 & 323 & 2,183 (1.13$\times$) & 1,020 & 1,384 & 2,224 \\
\hline
\end{tabular}
\caption{Latency per decision and peak GPU memory of the world model (WM), the policy, and both together at the budget of Table~\ref{tab:main}, on normal-speed Drop to Hoop with one NVIDIA RTX 5000 Ada GPU (means of two rounds of four measured episodes per method). Latency is per decision that reads the world model: ten world-model updates, one prediction, and the policy call with the one-time score amortized (hence the higher policy latency of ActPrune). The memory total is the peak of the simultaneous sum of the two processes, which can be below the sum of their separate peaks.}
\label{tab:latency-deploy}
\end{table}
}{}

\subsection{Why Does the Pruned Model Still Act Correctly?}
\label{sec:token-analysis}

The world model influences the action only through the dynamics tokens (Sec.~\ref{sec:background}), so we compare these tokens before and after pruning at the budget of Table~\ref{tab:main} with ActPrune, opacity, random selection, and farthest-point sampling on eight saved states (supplementary material, Sec.~B).
First, pruning changes the tokens substantially: every criterion loses 31--51\% of the motion signal, the part of a token that differs from the constant token of an empty neighborhood, and ActPrune loses 40\%.
Second, the action barely depends on the part that ActPrune removes: when each token is weighted by the sensitivity of the action to it, ActPrune keeps at least 99.5\% of the motion signal in every state, whereas opacity keeps 80\% and 58\% on the two Drop to Hoop variants and farthest-point sampling at most 28\%.
Third, the weighted signal is consistent with the closed-loop results of ActPrune, but it does not predict success in general: it is high for ActPrune partly by construction, since score and weights come from the same gradients, and a criterion that loses part of it does not always lose success.

\section{Conclusion}
\label{sec:conclusion}

We introduced ActPrune, which prunes the Gaussians of an online physical world model by how strongly the action of a frozen policy depends on them, and compared it with five existing criteria: opacity and four rendering-based ones.
First, rendering quality does not decide the success of the policy: with all robot Gaussians and 3\% of the others kept, the criteria that render the pruned scene best (\RQlg{} and \RQtaming\,dB PSNR) do not succeed most, and ActPrune, which renders it worst (\RQours\,dB), has the highest mean success rate over the 16 task variants of PhysMani-Bench (46.3\%, against 42.8\% for PUP 3D-GS and 42.3\% for opacity; on par with the full model, 44.7\%).
Second, where the policy relies on the world model, what is kept decides success: the best-rendering criteria keep almost none of the task objects, whereas ActPrune keeps a larger share of them than every other criterion on nearly every task. The criteria differ most on Drop to Hoop, where ActPrune keeps the few Gaussians of the small moving ball and succeeds in 50 of 80 trials, against 48 for the full model.
Third, at this budget the peak GPU memory of world model and policy falls from 3.6 to 2.2\,GiB and each decision is 1.13$\times$ faster.


\section*{LLM Usage Statement}
We used ChatGPT (OpenAI) as assistive tools on
content written or produced by the authors. Their use was limited to the
following parts of the work:
\begin{itemize}
  \item \textbf{Text:} editing author-written text for grammar, wording,
        and readability.
  \item \textbf{Related work and citations:} checking the authors'
        descriptions of cited works for accuracy, suggesting revisions to
        their wording, and pointing to related prior work. The authors read
        and verified every cited work before including it.
  \item \textbf{Code:} debugging and refining code written by the authors.
  \item \textbf{Experiments:} refining author-written evaluation and
        analysis scripts. The experiments were designed, run, and
        interpreted by the authors.
  \item \textbf{Figures:} refining the styling and layout of figures
        produced by the authors, such as editing plotting scripts.
\end{itemize}
No LLMs or generative AI tools were used to create or annotate data, or to
develop research ideas, hypotheses, methods, claims, or conclusions. The
authors reviewed and verified all AI-assisted content and take full
responsibility for the paper.

{
    \small
    \bibliographystyle{ieeenat_fullname}
    \bibliography{research}
}


\end{document}